\documentclass[letterpaper, 10pt, conference]{ieeeconf}      %

\IEEEoverridecommandlockouts                              %

\usepackage{amsmath,amsfonts,bm}

\def\eqref#1{Eq.~(\ref{#1})}

\def\1{\bm{1}}

\def\rvx{{\mathbf{x}}}
\def\rvy{{\mathbf{y}}}
\def\rvz{{\mathbf{z}}}

\def\vmu{{\bm{\mu}}}

\def\vchi{{\bm{\chi}}}

\def\mI{{\bm{I}}}

\def\mSigma{{\bm{\Sigma}}}

\DeclareMathAlphabet{\mathsfit}{\encodingdefault}{\sfdefault}{m}{sl}
\SetMathAlphabet{\mathsfit}{bold}{\encodingdefault}{\sfdefault}{bx}{n}

\newcommand{\E}{\mathbb{E}}

\newcommand{\KL}{D_{\mathrm{KL}}}

\usepackage{booktabs}       %
\usepackage{amsfonts}       %
\usepackage{multicol}
\usepackage{multirow}        %
\usepackage{xspace}
\usepackage{adjustbox}

\usepackage{amsmath} %
\usepackage{amssymb}  %
\usepackage{mathtools}
\usepackage{siunitx}

\usepackage[table,svgnames]{xcolor}
\usepackage{makecell}
\usepackage{graphicx}
\usepackage{tabularx,booktabs}
\newcolumntype{Y}{>{\centering\arraybackslash}X}
\newcolumntype{t}{>{\raggedleft\hsize=.03\hsize}X}
\newcolumntype{s}{>{\hsize=.4\hsize}X}

\usepackage[nolist]{acronym}
\usepackage{flushend}
\usepackage[font=small]{caption}
\usepackage[font=small]{subcaption}
\usepackage{mfirstuc} %
\MFUnocap{the}\MFUnocap{and}\MFUnocap{of}\MFUnocap{with}\MFUnocap{for}\MFUnocap{a}\MFUnocap{by}\MFUnocap{to}

\makeatletter
\renewcommand*\env@matrix[1][*\c@MaxMatrixCols c]{%
	\hskip -\arraycolsep
	\let\@ifnextchar\new@ifnextchar
	\array{#1}}
\makeatother

\makeatletter
\let\NAT@parse\undefined
\makeatother

\usepackage{tikz}
\usetikzlibrary{arrows}
\usetikzlibrary{fadings}
\usetikzlibrary{patterns}
\usetikzlibrary{shadows.blur}
\usetikzlibrary{shapes}
\tikzset{
	vertex/.style={circle,draw,minimum size=2.5em},
	edge/.style={->,> = latex'}
}

\usepackage{pifont}%

\usepackage{hyperref}

\begin{acronym}
\acro{GPU}{Graphics Processing Units}
\acro{DNN}{Deep Neural Networks}
\acro{CNN}{Convolutional Neural Network}
\acro{RNN}{Recurrent Neural Networks}
\acro{GNN}{Graph Neural Network}
\acro{GAT}{Graph Attention Network}
\acro{GCN}{Graph Convolutional Networks}
\acro{GRU}{Gated Recurrent Unit}
\acro{LSTM}{Long Short-Term Memory}
\acro{KF}{Kalman Filters}
\acro{BEV}{Bird's Eye View}
\acro{GAN}{Generative Adversarial Network}
\acro{MTP}{Multi-Modal Trajectory Prediction}
\acro{MLP}{Multilayer Perceptron}
\acro{MAE}{Mean Absolute Error}
\acro{MSE}{Mean Squared Error}
\acro{ADE}{Average Displacement Error}
\acro{FDE}{Final Displacement Error}
\acro{minADE}{Minimum Average Displacement Error}
\acro{minFDE}{Minimum Final Displacement Error}
\acro{DKM}{Deep Kinematic Models}
\acro{FF-ASP}{Feed-Forward Action-Space Predictor}
\acro{SS-ASP}{Self-Supervised Action-Space Predictor}
\acro{VAE}{Variational Autoencoders}
\acro{CVAE}{Conditional Variational Autoencoder}
\acro{AV}{Automated Vehicle}
\acro{MHA}{Multi-Head Attention}
\acro{SAN}{Scene Anchor Networks}
\acro{MB-SS-ASP}{Multi-Branch Self-Supervised Action-Space Predictor}
\acro{AD}{Automated Driving}
\acro{DL}{Deep Learning}
\acro{RL}{Reinforcement Learning}
\acro{MR}{Miss Rate}
\acro{IL}{Imitation Learning}
\acro{MDP}{Markov Decision Process}
\acro{POMDP}{Partially Observable Markov Decision Process}
\acro{RSSM}{Recurrent State Space Models}
\acro{VAE}{Variational Autoencoder}
\acro{UAE}{Unscented Autoencoder}
\acro{UT}{Unscented Transform}
\acro{CUAE}{Conditional Unscented Autoencoder}
\acro{GMM}{Gaussian Mixture Model}
\acro{NF}{Normalizing Flow}
\acro{KL}{Kullback-Leibler}
\acro{UT}{Unscented Transform}
\acro{FID}{Fréchet Inception Distance}
\acro{XP}{ex-post}
\acro{CXP}{conditional ex-post}
\acro{NLL}{Negative Log Likelihood}
\acro{ELBO}{Evidence Lower Bound}
\acro{WTA}{winner-takes-all}
\end{acronym}

\title{\LARGE \bf Conditional Unscented Autoencoders for Trajectory Prediction
}

\author{Faris Janjo\v{s}$^{1}$ Marcel Hallgarten$^{1,2}$ Anthony Knittel$^{1,3}$ Maxim Dolgov$^{1}$ Andreas Zell$^{2}$ J. Marius Z\"ollner$^{4}$%
\thanks{$^{1}$ Robert Bosch GmbH, Corporate Research, Renningen, Germany {\tt\small first-name.last-name@bosch.com}; $^{2}$ University of Tübingen, Tübingen, Germany; $^{3}$ Five AI Ltd, Cambridge, United Kingdom; $^{4}$ FZI Research Center for Information Technology, Karlsruhe, Germany}%
}
\begin{document}
\maketitle

\vspace{-10pt}
\begin{abstract}
The \ac{CVAE} is one of the most widely-used models in trajectory prediction for \ac{AD}. It captures the interplay between a driving context and its ground-truth future into a probabilistic latent space and uses it to produce predictions. In this paper, we challenge key components of the \ac{CVAE}. We leverage recent advances in the space of the \ac{VAE}, the foundation of the \ac{CVAE}, which show that a simple change in the sampling procedure can greatly benefit performance. We find that unscented sampling, which draws samples from any learned distribution in a deterministic manner, can naturally be better suited to trajectory prediction than potentially dangerous random sampling. We go further and offer additional improvements including a more structured Gaussian mixture latent space, as well as a novel, potentially more expressive way to do inference with \ac{CVAE}s. We show wide applicability of our models by evaluating them on the INTERACTION prediction dataset, outperforming the state of the art, as well as at the task of image modeling on the CelebA dataset, outperforming the baseline vanilla \ac{CVAE}. Code is available at: \footnotesize{\url{https://github.com/boschresearch/cuae-prediction}}.
\end{abstract}

\section{Introduction}\label{sec:intro}
Predicting the motion of human-driven vehicles sharing an environment with an autonomous system is a key enabler for fully-automated driving. Rich environment contexts present in urban driving and the prevalent interaction between traffic participants make it imperative to model the uncertainty in future trajectories. In this task of probabilistic trajectory prediction, machine learning models have proven indispensable. By learning a probability distribution, either in the space of the model's internal representations or the model's output, they capture the uncertainty inherent to the problem.%

In addressing the challenges of probabilistic trajectory prediction, many approaches use established generative models such as a \ac{CVAE}, a \ac{GAN}, or a \ac{NF}. The \ac{CVAE} is especially useful; its powerful latent space model represents the underlying structure present in the relationship between a future trajectory and the potentially high-dimensional historical context that induces it. This real-world joint distribution is compressed into a tractable, relatively low-dimensional latent space Gaussian, amenable to sampling. Generating future predictions involves simply drawing samples from the latent space and transforming them into trajectories. The tasks of compressing inputs into the latent space and decompressing predictions from it are delegated to the \ac{CVAE}'s encoder and decoder, which can leverage powerful GNN or Transformer models. Thus, it has been a method of choice in state-of-the-art probabilistic prediction~\cite{ivanovic2020multimodal,salzmann2020trajectron++,casas2020implicit,cui2021lookout,yuan2021agentformer,lee2022muse}.%

\begin{figure}[t]
    \centering
    \hspace{-20pt}
    \scalebox{0.45}{\input{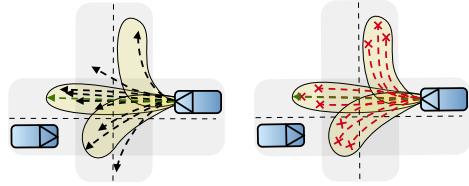}}
    \caption{\footnotesize{Assume a trajectory predictor learned a multi-modal distribution (yellow), either by propagating its latent space or directly in the output space. Random sampling (black) can bring unsafe, unlikely, or in-between-mode outputs. In contrast, the unscented sampling (red), realized by computing sigma points of the distribution, brings structure to the learned stochasticity.}}
    \label{fig:sampling-illustration}
    \vspace{-20pt}
\end{figure}

Despite its wide appeal, the \ac{CVAE} has certain shortcomings when applied in trajectory prediction. It does not provide an out-of-the-box means to evaluate the likelihoods of its trajectories. Further, since the distribution of future motion is highly multi-modal (usually involving distinct behaviors), recovering it from a smooth Gaussian latent space used in continuous \ac{CVAE}s can bring unreasonable in-between outputs. Finally, the randomness inherent to the model is at odds with the primacy of safety and reproducibility. Random sampling of the latent space in inference can generate spurious and potentially dangerous trajectories (see Fig.~\ref{fig:sampling-illustration}) as well as miss critical trajectories, in addition to bringing a high gradient variance in training (a pitfall of the \ac{VAE} itself). This can have serious ramifications on a downstream planner fed \ac{CVAE}-predicted futures; these might differ significantly over consecutive prediction calls. Overall, these issues can be traced back to the \ac{CVAE}'s overly simplistic latent space and the unreliable random sampling.

Our work challenges well-established assumptions surrounding the \ac{CVAE} and its Gaussian latent space. We aim to answer the following two research questions: \hypertarget{rq1}{\textit{(i) Can the random sampling and propagation be replaced by more structured selection?}}, \hypertarget{rq2}{\textit{(ii) Are there effective alternatives to the simplistic latent space in training and inference, especially considering multi-modality of the output space?}} In answering \hyperlink{rq1}{\textit{(i)}}, we leverage recent advancements in the base \ac{VAE}~\cite{janjovs2023unscented} while for \hyperlink{rq2}{\textit{(ii)}}, we use more expressive distributions. As we improve core aspects of the \ac{CVAE}, we also evaluate our models on image generation tasks. Our contributions are:
\begin{itemize}
\item Unscented sampling and transformation of \ac{CVAE} distributions as an alternative to random sampling for trajectory prediction, tackling \hyperlink{rq1}{\textit{(i)}}. As part of this contribution, we develop a novel \ac{CUAE} model with deterministic sampling.
\item A \ac{CVAE} extension toward a mixture model latent space in place of a Gaussian latent space (usable with both random and unscented sampling). It promotes multi-modality in the output-space and tackles \hyperlink{rq2}{\textit{(ii)}}.
\item A novel approach for inference with \ac{CVAE}s via conditional ex-post estimation, inspired by~\cite{janjovs2023unscented} and tackling \hyperlink{rq2}{\textit{(ii)}}. It preserves latent space training but circumvents the need to use it in inference by building and sampling a more expressive distribution instead.
\end{itemize}

\section{Related Work}
In the following, we discuss approaches to probabilistic trajectory prediction, i.e. modeling the conditional distribution $\mathcal{P}(\rvy|\rvx)$ of future trajectories $\rvy$ given a generic context $\rvx$.
A popular choice is to represent $\mathcal{P}(\rvy|\rvx)$ by a \textbf{set of trajectories}. Here, a fixed number of modes with associated probabilities is regressed either individually per agent~\cite{gao2020vectornet,cui2020deep,gilles2021home,zhao2021tnt,gilles2022gohome} or jointly for all agents in a scene~\cite{liang2020learning,gilles2021thomas,ngiam2021scene,janjovs2022starnet,cui2023gorela}. Commonly, \ac{WTA} loss functions only consider the predicted mode closest to the ground truth, exhibiting low sample efficiency. Moreover, as non-winner modes are not penalized, the predicted set can contain unrealistic and inadmissible trajectories (e.g. off-road). Many approaches address such issues by explicitly conditioning on map elements~\cite{narayanan2021divide,zhao2021tnt,gu2021densetnt,deo2022multimodal,hallgarten2023stay}.

Other classes of models attempt to directly capture $\mathcal{P}(\rvy|\rvx)$ into a \textbf{parametric distribution} such as a \ac{GMM}. Here, trajectories are considered means of mixture components and (co)variances are learned separately~\cite{chai2019multipath,khandelwal2020if,phan2020covernet,varadarajan2022multipath++,knittel2023dipa}. This approach models the uncertainty of the underlying problem more accurately. Moreover, loss functions can consider the entire distribution (via the \ac{NLL}), increasing sample efficiency and reducing inadmissible predictions. However, these models are theoretically limited since they do not reason about the generative process of the data, i.e. $\mathcal{P}(\rvx, \rvy)$.

In contrast, \textbf{generative models} such as \ac{NF}s~\cite{rhinehart2019precog,scholler2021flomo,meszaros2023trajflow}, \ac{GAN}s~\cite{gupta2018social,sadeghian2019sophie,kosaraju2019social,goodfellow2020generative,dendorfer2020goal}, or \ac{CVAE}s~\cite{ivanovic2020multimodal,salzmann2020trajectron++,casas2020implicit,cui2021lookout,yuan2021agentformer,lee2022muse},
attempt to first learn a proxy for the joint data distribution and then obtain the predictive distribution $\mathcal{P}(\rvy|\rvx)$. By sampling a prior and propagating the samples into the output space, they can implicitly capture rich non-parametric distributions. Among these models, \ac{NF}s have limited expressiveness for high-dimensional data distributions found in trajectory prediction as well as strict architectural constraints, although they provide tractable likelihoods. Among \ac{GAN}s, prevalent issues include lack of diversity and mode collapse~\cite{salimans2016improved}, out-of-distribution samples~\cite{tanielian2020learning,khayatkhoei2018disconnected}, and training instability~\cite{arjovsky2017towards,arjovsky2017wasserstein}. %
Moreover, \ac{GAN}s learn a continuous transformation and are unable to model disconnected manifolds~\cite{tanielian2020learning,khayatkhoei2018disconnected,dendorfer2021mg}, which is often necessary in prediction. To mitigate this,~\cite{dendorfer2021mg} uses multiple generators.

Similarly, \ac{CVAE}s can struggle to model output distributions with disconnected modes~\cite{rolfe2016discrete}. The decoder transformation is continuous and the latent distribution capturing multiple futures is commonly modeled as a multivariate Gaussian. The approaches in~\cite{salzmann2020trajectron++,hong2019rules} address the issue by using a discrete latent variable mapped to a \ac{GMM} output, which also facilitates integration over the conditional prior distribution. However, this limits the expressiveness of the latent space. In this work, we approach this problem by leveraging more expressive latent distributions such as \ac{GMM}s in both training and inference. Furthermore, many models adapt the \ac{CVAE} to output likelihoods by additional classifier networks, a common approach across the trajectory prediction landscape~\cite{casas2020implicit,cui2021lookout,knittel2023dipa}.
Another pitfall of \ac{CVAE}s is that propagating the latent distribution to the output space involves drawing and decoding random samples. The randomness can result in bad coverage of the true output distribution, especially with few samples. To mitigate this,~\cite{cui2021lookout} and~\cite{yuan2021agentformer} employ diversity sampling techniques in training, while~\cite{cui2021lookout} off-loads the modeling of distinct futures to a GNN decoder. In inference,~\cite{cui2021lookout} uses only the latent mean and abandons the rich learned latent space. In contrast, we use deterministic sampling~\cite{janjovs2023unscented} to obtain diverse and representative samples from the learned latent space.

\section{Method}\label{sec:method}
We consider the task of modeling $\mathcal{P}(\rvy|\rvx)$, where $\rvy$ are future vehicle trajectories $\rvy$, i.e. a $T{\times}2$ matrix of $T$ future positions, and $\rvx$ is a generic context. In addressing this task, we leverage the \ac{CVAE} framework to construct and sample an expressive latent space. Thus, we outline the presentation of our approach along the research questions posed in Sec.~\ref{sec:intro}. Sec.~\ref{subsec:lat_sampling} provides a \ac{CVAE} background and proposes alternatives to random sampling and transformation of the latent space, tackling \hyperlink{rq1}{\textit{(i)}}, while Sec.~\ref{subsec:lat_representation} offers alternatives in latent space modeling and using it for inference, tackling \hyperlink{rq2}{\textit{(ii)}}. Sec.~\ref{subsec:traj_gen} discusses the generation of output trajectories given the choices in Sec.~\ref{subsec:lat_sampling} and Sec.~\ref{subsec:lat_representation}.
\subsection{Latent Space Sampling and Transformation}\label{subsec:lat_sampling}
\subsubsection{\ac{CVAE} Background}\label{subsubsec:cvae}
\begin{figure}
    \centering
    \scalebox{0.64}{\input{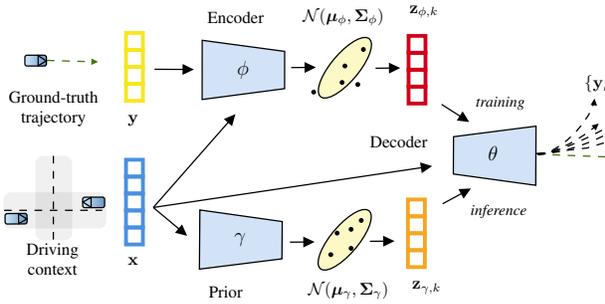}}
    \caption{\footnotesize{CVAE: in training, the model captures the joint distribution of the ground-truth trajectory and driving context via the encoder network $\phi$, samples randomly, and reconstructs trajectories via the decoder network $\theta$. In inference, the prior network $\gamma$ replaces $\phi$ and is sampled instead.}}
    \label{fig:cvae}
    \vspace{-15pt}
\end{figure}

\ac{CVAE}s ~\cite{sohn2015learning} are generative models that can capture a conditional distribution $\mathcal{P}(\rvy|\rvx)$. They model the relationship between pairs of high-dimensional inputs $\rvx$ and $\rvy$ by projecting them into a lower-dimensional latent space $\rvz$, see Fig.~\ref{fig:cvae}. An encoder parameterized by $\phi$ learns the latent posterior distribution $q_{\phi}(\rvz|\rvx, \rvy; \vmu_\phi, \mSigma_\phi)$, commonly modeled as a multivariate Gaussian. Then, a $\theta$-parameterized decoder is tasked with estimating the true output distribution $\mathcal{P}(\rvy|\rvx)$. This is done by conditioning on $\rvx$ and drawing random samples $\rvz$ from $q_{\phi}$ to first produce $p_{\theta}(\rvy|\rvx,\rvz)$ and thus marginalize out $\rvz$. In inference, since the ground-truth $\rvy$ is not available, the model instead samples a surrogate, $\gamma$-parameterized latent prior $p_{\gamma}(\rvz|\rvx;\vmu_\gamma, \mSigma_\gamma)$. The posterior $q_\phi$ and prior $p_\gamma$ are trained to be consistent. Thus, the loss function minimizes $\mathcal{L}_{\text{CVAE}}=\mathcal{L}_{\text{REC}}+\mathcal{L}_{\text{KL}}$ (and maximizes the \ac{ELBO}~\cite{kingma2013auto}),
\begin{align}
\mathcal{L}_{\text{REC}} &= -\E_{\rvz\sim q_\phi(\rvz|\rvx,\rvy)}\left[\log p_\theta (\rvy|\rvx,\rvz)\right]\label{eq:cvae_rec}\ ,\\ 
\mathcal{L}_{\text{KL}} &= \KL(q_\phi(\rvz|\rvx,\rvy)\Vert p_\gamma(\rvz|\rvx))\label{eq:cvae_kl}\ .
\end{align}

The term in~\eqref{eq:cvae_rec} promotes consistency between the decoder output and the observed ground truth, while~\eqref{eq:cvae_kl} brings the posterior and prior distributions together by minimizing their \ac{KL} divergence. In practice, $K$ samples $\rvz_k$ from $q_\phi$ are drawn and a deterministic decoder function $f_{\theta}$ maps each ($\rvx, \rvz_k$) pair to an output trajectory $\rvy_k{=}f_{\theta}(\rvx,\rvz_k)$. Thus, no parametric form of $\mathcal{P}(\rvy|\rvx)$ is estimated and the output distribution is represented by a set of $K$ samples. In this way, the reconstruction term in~\eqref{eq:cvae_rec} can be approximated by the \ac{NLL}\footnote{In image modeling, it is usually the \ac{MSE} instead.} of reconstructed samples $\rvy_k$ under the ground-truth distribution $\mathcal{N}(\rvy, \sigma\mI)$, yielding
\begin{equation}\label{eq:rec-samples}
\mathcal{L}_{\text{REC-samples}} = -\textstyle\frac{1}{K}\sum^K_k\log\mathcal{N}(\rvy_k; \rvy, \sigma\mI)\ . 
\end{equation}

A common approximation when predicting entire trajectories in one shot is to assume independence across time steps~\cite{knittel2023dipa} or a fixed diagonal covariance matrix $\sigma\mI$~\cite{dendorfer2021mg}. 

\subsubsection{Unscented Transform of the Latent Space}\label{subsubsec:cuae}
A key component of the \ac{CVAE} (and its \ac{VAE} foundation) is random sampling of the latent space. It is a feature of the reparameterization trick~\cite{kingma2013auto}, employed in order to sample the latent Gaussian posterior and efficiently compute gradients w.r.t. $\phi$ in ~\eqref{eq:cvae_rec}. However, it exhibits high variance in training. Therefore, a deterministic-sampling alternative based on the \ac{UT}~\cite{julier2004unscented} (prominent in filtering and control) has emerged in the \ac{UAE}~\cite{janjovs2023unscented}. It is motivated by the fact that the decoder is a nonlinear function of the posterior distribution. Thus, a set of representative points in the latent space can be chosen and transformed to approximate the output distribution, which is difficult in practice by transforming a few random samples. 

The \ac{UT} application in the \ac{VAE} context can be straightforwardly extended to the \ac{CVAE}. The \ac{CUAE} model is shown in Fig.~\ref{fig:cuae}. The model analytically computes the sigma points of the Gaussian posterior $\mathcal{N}(\vmu_\phi, \mSigma_\phi)$, $\vmu_\phi\in\mathbb{R}^n$. The $2n{+}1$ sigmas $\{\vchi_i\}^{2n}_{i=0}$ are the mean $\vchi_n{=}\vmu_\phi$ and a pair on each axis $\vchi_{n\pm j}{=}\vmu_\phi{\pm}\sqrt{n\mSigma_\phi} \big{|}_j\ ,1{\leq}j{\leq}n$, where $\big{|}_j$ indicates the $j$-th column. For a commonly used diagonal $\mSigma_\phi$, there is no computational overhead since the Cholesky decomposition $\sqrt{\mSigma_\phi}$ can be directly obtained from predicted log variances. 

Since the sigmas fully describe the latent distribution\footnote{Their first two moments (the  mean and covariance) equal the original distribution's first two moments.}, they usually also describe the output distribution well w.r.t commonly used decoder nonlinearities~\cite{julier2000new}. Thus, going one step further, we can approximate the expectation in~\eqref{eq:cvae_rec} by the mean of the transformed sigmas. With this, we push the entire output distribution (w.r.t. its mean) toward the ground truth instead of the individually transformed samples. Thus,
\begin{equation}\label{eq:rec-dist}
\mathcal{L}_{\text{REC-dist.}} = -\log\mathcal{N}(\textstyle\frac{1}{K}\sum^K_k\rvy_k; \rvy, \sigma\mI)\ ,
\end{equation}
where each $\rvy_k$ comes from a latent space sigma point. In practice, due to a large dimensionality $n$, we select $K{<}2n{+}1$ random pairs of sigma points on the same covariance axis~\cite{janjovs2023unscented}.

In the context of trajectory prediction, sigma points have the potential to reasonably cover the latent space with few samples, train the entire output distribution accordingly, and prevent spurious and unlikely samples in inference. %
\begin{figure}
\captionbox{\label{fig:cuae} \footnotesize{\ac{CUAE}: instead of sampling the latent space randomly (in both training and inference), the model analytically computes sigma points of the $\phi$ and $\gamma$ distributions and transforms them instead.}}[{0.47\columnwidth}]{\scalebox{0.61}{\input{figures/cuae_decoder.tex}}}
\hspace{0.01cm}
\captionbox{\label{fig:gmm_cuae} \footnotesize{\ac{GMM}-\ac{CUAE}: it structures the latent space into a \ac{GMM} and separately transforms its components (sigma points shown). Compared to Fig.~\ref{fig:cuae}, it has the potential to better model multi-modality.}}[{0.47\columnwidth}]{\scalebox{0.61}{\input{figures/gmm_cuae_decoder.tex}}}
\vspace{-15pt}
\end{figure}

\vspace{-5pt}
\subsection{Latent Space Structure and Inference Strategy}\label{subsec:lat_representation}
Many use-cases within probabilistic trajectory prediction necessitate a disjoint output with well-separated modes such as turning left or right. Here, \ac{CVAE}s struggle due to the continuous latent distribution that is decoded as a continuous distribution of trajectories. Therefore, we propose two methods to promote a multi-modal output space. Both use a \ac{GMM} structure: the first imposes it to the latent space and the second uses a separately-constructed \ac{GMM} purely for inference.\looseness=-1

\subsubsection{GMM Latent Space}\label{subsubsec:mixture_cvae}
The mixture prior model attempts to capture distinct modes of behavior using a \ac{GMM} for the prior and posterior distributions in the latent space, see Fig.~\ref{fig:gmm_cuae}. The GMM components can correspond with modes of behavior, while the distribution of each can represent the variation within each mode. For example, one mode may correspond to a right-turn behavior whose speed or path variation is given by the variance.  %
The two GMMs with $C$ components are described by $\sum^C_c w_\phi^{(c)} \mathcal{N}(\vmu_\phi^{(c)}, \mSigma_\phi^{(c)})$ and $\sum^C_c w_\gamma^{(c)} \mathcal{N}(\vmu_\gamma^{(c)}, \mSigma_\gamma^{(c)})$, for fixed $C$. Sampling is performed independently for each component; we draw $K$ random samples or sigma points from each, totaling $K{\cdot}C$. %
Then, we compute the centroid $\rvy^{(c)}_{\vmu}$ and covariance $\rvy^{(c)}_{\mSigma}$ of the decoded trajectories for each mode to obtain an output-space GMM, with the corresponding component weights carried over.\looseness=-1

The loss functions in \eqref{eq:cvae_rec} and \eqref{eq:cvae_kl} are adapted as follows to be compatible with a GMM representation. As the \ac{KL} divergence between Gaussian mixtures is not analytically defined, we apply it individually (according to~\eqref{eq:cvae_kl}) between each corresponding posterior and prior component and their discrete mixture distributions %
\begin{equation}
\label{eq:mixture_kl}
\mathcal{L}_{\text{KL-GMM}} = \textstyle\sum_c^C
\mathcal{L}^{(c)}_{\text{KL}} 
+ \KL(w_{\phi}\Vert w_{\gamma})\ .
\end{equation}
The reconstruction loss minimizes the \ac{NLL} of the ground-truth trajectory under the predicted future distribution, represented by the output GMM.
This way, we train the component whose centroid trajectory is closest to the ground-truth $\rvy$ (denoted by $c^*$) and a corresponding one-hot distribution $w_\rvy$ of the closest component, 
\vspace{-1pt}
\begin{equation}
	\mathcal{L}_{\text{REC-GMM}} = -%
 \log \mathcal{N}(\rvy; \rvy^{(c^*)}_{\vmu}, \rvy^{(c^*)}_{\mSigma}) 
    + \KL(w_\phi\Vert w_\rvy)\ . \label{eq:covar_mm_reconst}
\vspace{-1pt}
\end{equation}
Note that this form of the reconstruction loss function is closer to the \ac{UT}-like computation in~\eqref{eq:rec-dist} than the vanilla \ac{CVAE}-like computation in~\eqref{eq:rec-samples}, since the likelihood is computed for the centroid trajectory of the winner latent \ac{GMM} component rather than in expectation over multiple individual latent samples. The choice between random and unscented sampling determines the method to obtain the output trajectories (that approximate the true distribution), which are averaged to compute the centroid.

\subsubsection{Conditional Ex-Post (CXP) Estimation}\label{subsubsec:cond-expost}
\begin{figure}[t!]
\begin{subfigure}{1\textwidth}
\scalebox{0.6}{\input{figures/cond-expost-train.tex}}\hspace{-0.37cm}
\includegraphics[width=0.13\textwidth]{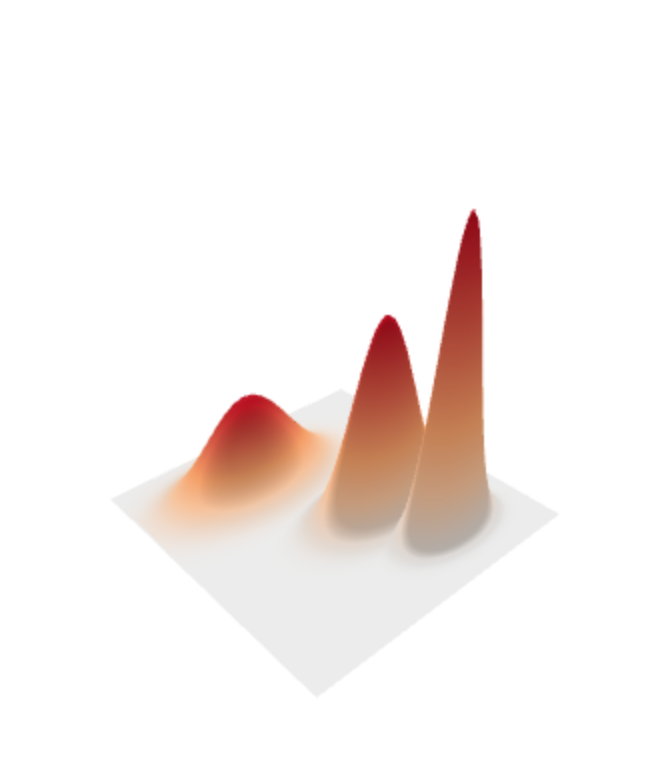}\hspace{0.37cm}
\end{subfigure}
\begin{subfigure}{1\textwidth}
\scalebox{0.6}{\tikzset {_ez2x8lyt6/.code = {\pgfsetadditionalshadetransform{ \pgftransformshift{\pgfpoint{0 bp } { 0 bp }  }  \pgftransformrotate{0 }  \pgftransformscale{2 }  }}}
\pgfdeclarehorizontalshading{_2s6ip39x1}{150bp}{rgb(0bp)=(0.81,0.91,0.98);
rgb(37.5bp)=(0.81,0.91,0.98);
rgb(62.5bp)=(0.39,0.58,0.76);
rgb(100bp)=(0.39,0.58,0.76)}

\tikzset {_wbn6de9uc/.code = {\pgfsetadditionalshadetransform{ \pgftransformshift{\pgfpoint{0 bp } { 0 bp }  }  \pgftransformrotate{0 }  \pgftransformscale{2 }  }}}
\pgfdeclarehorizontalshading{_1qblrsrz8}{150bp}{rgb(0bp)=(0.81,0.91,0.98);
rgb(37.5bp)=(0.81,0.91,0.98);
rgb(62.5bp)=(0.39,0.58,0.76);
rgb(100bp)=(0.39,0.58,0.76)}
\tikzset{every picture/.style={line width=0.75pt}} %

\begin{tikzpicture}[x=0.75pt,y=0.75pt,yscale=-1,xscale=1]

\draw  [color={rgb, 255:red, 245; green, 166; blue, 35 }  ,draw opacity=1 ][line width=2.25]  (164.37,58.1) -- (177.7,58.1) -- (177.7,71.43) -- (164.37,71.43) -- cycle ;
\draw  [color={rgb, 255:red, 245; green, 166; blue, 35 }  ,draw opacity=1 ][line width=2.25]  (164.37,71.43) -- (177.7,71.43) -- (177.7,84.77) -- (164.37,84.77) -- cycle ;
\draw  [color={rgb, 255:red, 245; green, 166; blue, 35 }  ,draw opacity=1 ][line width=2.25]  (164.37,84.77) -- (177.7,84.77) -- (177.7,98.1) -- (164.37,98.1) -- cycle ;
\draw  [color={rgb, 255:red, 245; green, 166; blue, 35 }  ,draw opacity=1 ][line width=2.25]  (164.37,98.1) -- (177.7,98.1) -- (177.7,111.43) -- (164.37,111.43) -- cycle ;
\draw    (137.37,87.7) -- (153.84,87.56) ;
\draw [shift={(156.84,87.53)}, rotate = 179.51] [fill={rgb, 255:red, 0; green, 0; blue, 0 }  ][line width=0.08]  [draw opacity=0] (8.93,-4.29) -- (0,0) -- (8.93,4.29) -- cycle    ;
\draw  [color={rgb, 255:red, 245; green, 166; blue, 35 }  ,draw opacity=1 ][line width=2.25]  (378.37,58.1) -- (391.7,58.1) -- (391.7,71.43) -- (378.37,71.43) -- cycle ;
\draw  [color={rgb, 255:red, 245; green, 166; blue, 35 }  ,draw opacity=1 ][line width=2.25]  (378.37,71.43) -- (391.7,71.43) -- (391.7,84.77) -- (378.37,84.77) -- cycle ;
\draw  [color={rgb, 255:red, 245; green, 166; blue, 35 }  ,draw opacity=1 ][line width=2.25]  (378.37,84.77) -- (391.7,84.77) -- (391.7,98.1) -- (378.37,98.1) -- cycle ;
\draw  [color={rgb, 255:red, 245; green, 166; blue, 35 }  ,draw opacity=1 ][line width=2.25]  (378.37,98.1) -- (391.7,98.1) -- (391.7,111.43) -- (378.37,111.43) -- cycle ;
\draw    (398.37,88.6) -- (414.84,88.46) ;
\draw [shift={(417.84,88.43)}, rotate = 179.51] [fill={rgb, 255:red, 0; green, 0; blue, 0 }  ][line width=0.08]  [draw opacity=0] (8.93,-4.29) -- (0,0) -- (8.93,4.29) -- cycle    ;
\draw [color={rgb, 255:red, 0; green, 0; blue, 0 }  ,draw opacity=1 ][line width=0.75]  [dash pattern={on 4.5pt off 4.5pt}]  (476.14,87.98) .. controls (483.64,87.62) and (493.21,82.57) .. (501.84,74.5) ;
\draw [shift={(503.92,72.48)}, rotate = 134.69] [fill={rgb, 255:red, 0; green, 0; blue, 0 }  ,fill opacity=1 ][line width=0.08]  [draw opacity=0] (5.36,-2.57) -- (0,0) -- (5.36,2.57) -- cycle    ;
\draw [color={rgb, 255:red, 0; green, 0; blue, 0 }  ,draw opacity=1 ][line width=0.75]  [dash pattern={on 4.5pt off 4.5pt}]  (476.78,88.28) .. controls (485.75,88.01) and (498.62,73.38) .. (501.49,62.59) ;
\draw [shift={(502.02,59.7)}, rotate = 95.16] [fill={rgb, 255:red, 0; green, 0; blue, 0 }  ,fill opacity=1 ][line width=0.08]  [draw opacity=0] (5.36,-2.57) -- (0,0) -- (5.36,2.57) -- cycle    ;
\draw [color={rgb, 255:red, 0; green, 0; blue, 0 }  ,draw opacity=1 ][line width=0.75]  [dash pattern={on 4.5pt off 4.5pt}]  (476.62,88.25) .. controls (485.69,87.97) and (497.27,84.07) .. (507.7,77.84) ;
\draw [shift={(510.21,76.28)}, rotate = 147.16] [fill={rgb, 255:red, 0; green, 0; blue, 0 }  ,fill opacity=1 ][line width=0.08]  [draw opacity=0] (5.36,-2.57) -- (0,0) -- (5.36,2.57) -- cycle    ;
\draw [color={rgb, 255:red, 0; green, 0; blue, 0 }  ,draw opacity=1 ][line width=0.75]  [dash pattern={on 4.5pt off 4.5pt}]  (476.99,88.3) .. controls (485.82,88.03) and (505.42,76.43) .. (513.63,70.38) ;
\draw [shift={(515.92,68.59)}, rotate = 139.6] [fill={rgb, 255:red, 0; green, 0; blue, 0 }  ,fill opacity=1 ][line width=0.08]  [draw opacity=0] (5.36,-2.57) -- (0,0) -- (5.36,2.57) -- cycle    ;
\draw [color={rgb, 255:red, 0; green, 0; blue, 0 }  ,draw opacity=1 ][line width=0.75]  [dash pattern={on 4.5pt off 4.5pt}]  (476.71,87.98) -- (499.25,85.05) -- (515.07,79.54) ;
\draw [shift={(517.9,78.55)}, rotate = 160.78] [fill={rgb, 255:red, 0; green, 0; blue, 0 }  ,fill opacity=1 ][line width=0.08]  [draw opacity=0] (5.36,-2.57) -- (0,0) -- (5.36,2.57) -- cycle    ;
\draw  [color={rgb, 255:red, 0; green, 0; blue, 0 }  ,draw opacity=1 ][fill={rgb, 255:red, 74; green, 144; blue, 226 }  ,fill opacity=0.25 ] (83.17,61.71) -- (133.92,70.42) -- (133.87,99.81) -- (83.1,108.38) -- cycle ;

\draw  [draw opacity=0][fill={rgb, 255:red, 0; green, 0; blue, 0 }  ,fill opacity=0.06 ] (5.32,101.84) .. controls (5.32,99.34) and (7.34,97.3) .. (9.84,97.3) -- (64.22,97.19) .. controls (66.72,97.18) and (68.76,99.2) .. (68.76,101.71) -- (68.79,115.29) .. controls (68.8,117.79) and (66.77,119.83) .. (64.27,119.83) -- (9.89,119.95) .. controls (7.39,119.95) and (5.36,117.93) .. (5.35,115.43) -- cycle ;
\draw  [draw opacity=0][fill={rgb, 255:red, 0; green, 0; blue, 0 }  ,fill opacity=0.06 ] (43.1,74.5) .. controls (45.6,74.5) and (47.63,76.52) .. (47.63,79.02) -- (47.68,122.61) .. controls (47.68,125.11) and (45.66,127.15) .. (43.15,127.15) -- (29.56,127.16) .. controls (27.06,127.16) and (25.03,125.14) .. (25.03,122.64) -- (24.98,79.05) .. controls (24.98,76.55) and (27.01,74.52) .. (29.51,74.51) -- cycle ;
\draw  [dash pattern={on 4.5pt off 4.5pt}]  (5.67,108.55) -- (66.21,108.52) ;
\draw  [dash pattern={on 4.5pt off 4.5pt}]  (36.07,126.75) -- (36.07,74.27) ;
\path  [shading=_2s6ip39x1,_ez2x8lyt6] (6.29,111.89) .. controls (6.29,111.17) and (6.86,110.58) .. (7.58,110.57) -- (18.5,110.47) .. controls (19.22,110.46) and (19.81,111.04) .. (19.81,111.76) -- (19.85,115.67) .. controls (19.86,116.39) and (19.28,116.98) .. (18.56,116.99) -- (7.65,117.1) .. controls (6.93,117.1) and (6.34,116.52) .. (6.33,115.8) -- cycle ; %
 \draw  [line width=0.75]  (6.29,111.89) .. controls (6.29,111.17) and (6.86,110.58) .. (7.58,110.57) -- (18.5,110.47) .. controls (19.22,110.46) and (19.81,111.04) .. (19.81,111.76) -- (19.85,115.67) .. controls (19.86,116.39) and (19.28,116.98) .. (18.56,116.99) -- (7.65,117.1) .. controls (6.93,117.1) and (6.34,116.52) .. (6.33,115.8) -- cycle ; %

\draw  [line width=0.75]  (19.74,113.94) -- (14.67,116.83) -- (14.59,110.7) -- cycle ;
\path  [shading=_1qblrsrz8,_wbn6de9uc] (67.68,105.54) .. controls (67.68,106.26) and (67.1,106.85) .. (66.38,106.85) -- (55.47,106.88) .. controls (54.75,106.89) and (54.16,106.3) .. (54.16,105.58) -- (54.15,101.67) .. controls (54.15,100.95) and (54.73,100.36) .. (55.45,100.36) -- (66.36,100.33) .. controls (67.08,100.33) and (67.67,100.91) .. (67.67,101.63) -- cycle ; %
 \draw  [line width=0.75]  (67.68,105.54) .. controls (67.68,106.26) and (67.1,106.85) .. (66.38,106.85) -- (55.47,106.88) .. controls (54.75,106.89) and (54.16,106.3) .. (54.16,105.58) -- (54.15,101.67) .. controls (54.15,100.95) and (54.73,100.36) .. (55.45,100.36) -- (66.36,100.33) .. controls (67.08,100.33) and (67.67,100.91) .. (67.67,101.63) -- cycle ; %

\draw  [line width=0.75]  (54.25,103.4) -- (59.33,100.55) -- (59.38,106.68) -- cycle ;

\draw  [color={rgb, 255:red, 0; green, 0; blue, 0 }  ,draw opacity=1 ][fill={rgb, 255:red, 74; green, 144; blue, 226 }  ,fill opacity=0.25 ] (472.37,111.44) -- (421.62,102.75) -- (421.65,73.36) -- (472.42,64.77) -- cycle ;

\draw (163.37,126.17) node [anchor=north west][inner sep=0.75pt]  [font=\normalsize]  {$\mathbf{z}_{\gamma ,k}$};
\draw (377.37,126.17) node [anchor=north west][inner sep=0.75pt]  [font=\normalsize]  {$\mathbf{z}_{cond-mix}$};
\draw (438.4,81.6) node [anchor=north west][inner sep=0.75pt]  [font=\large]  {$\theta $};
\draw (101.9,78.03) node [anchor=north west][inner sep=0.75pt]  [font=\large]  {$\gamma $};
\draw (7.93,34.23) node [anchor=north west][inner sep=0.75pt]  [font=\small] [align=left] {testing\\examples};

\end{tikzpicture}}\hspace{-5.6cm}
\includegraphics[width=0.175\textwidth,trim={0 0 0 6.5cm},clip]{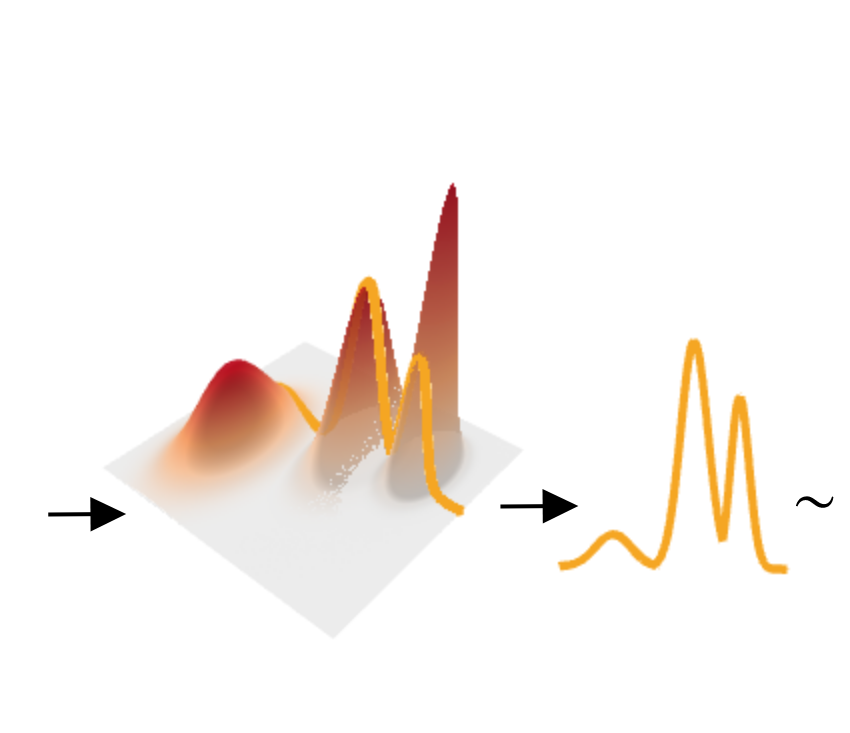}\hspace{5.6cm}
\end{subfigure}
\caption{\footnotesize{Illustration of \ac{CXP} joint mixture construction and conditional sampling. Top: all posterior and prior sigma points in training are concatenated, collected, and used to fit a mixture. Bottom: given a new example's prior encoding, the mixture is conditioned (intuitively, it is ``cut"). The resulting lower-dim. mixture is sampled as input for the decoder.}}
\label{fig:cond-expost-sampling}
\vspace{-15pt}
\end{figure}

In this section, we present an alternative to using the trained latent space for inference, which is universal among \ac{VAE}s and \ac{CVAE}s. Termed \ac{XP} estimation, it involves training a latent space but not using it directly in inference. For the \ac{VAE}, an empirically-obtained distribution constructed after training is sampled instead of the theoretically-imposed standard normal prior~\cite{ghosh2020variational}. The advantage is that it addresses the well-known \ac{VAE} posterior mismatch\footnote{In practice, the average posterior over the entire training set does not fully match the assumed $\mathcal{N}(\mathbf{0}, \mathbf{I})$ prior, leading to lower sample quality.} as well as enables deterministic autoencoders (lacking a probabilistic latent space) to generate samples. It can be realized by collecting a dataset of posterior encodings during training $\mathcal{D}_{\text{ex-post}}{=}\{\rvz^i_{\phi}\}$ ($i$ for training example) and using an off-the-shelf tool (e.g.~\cite{scikit-learn}) to fit another more expressive distribution. For example, a \ac{GMM} with $C$ Gaussians, $p(\rvz_{\phi\text{mix}}){=}\textstyle\sum^C_c w^{(c)} \mathcal{N}(\rvz_{\phi\text{mix}}; \vmu_{\phi\text{mix}}, \mSigma_{\phi\text{mix}})$, where $\rvz_{\phi\text{mix}}$ is the \ac{GMM} random variable, $w^{(c)}$ are the obtained weights, and $C$ is a priori defined. This empirical distribution is then sampled instead.\looseness=-1

We extend the original method to the \ac{CVAE} and \ac{CUAE} by introducing a conditional ex-post (CXP) estimated density. The vanilla \ac{XP} sampling is inadequate in the \ac{CVAE} case since using a mixture built only from the posterior encodings $\rvz_{\phi}$ precludes conditioning, e.g. the driving context encountered in a test set example. Therefore, we incorporate the prior encoding $\rvz_{\gamma}$ (obtained through the conditional prior $\gamma$, see Fig.~\ref{fig:cvae}). First, we collect a set of concatenated posterior-prior pairs $\mathcal{D}_{\text{cond-ex-post}}{=}\{[\rvz^i_{\phi}; \rvz^i_{\gamma}]\}$ and then fit a \ac{GMM} $p(\rvz_{\phi\gamma\text{mix}})$. %
See Fig.~\ref{fig:cond-expost-sampling} (top) for an illustration. These pairs are a dataset of latent-space relationships between future trajectories and the associated context. Thus, concatenating them and fitting a \ac{GMM} models the joint distribution between the ground-truth-future-posterior and the context-prior that preceded it. %
However, sampling given a new context $\rvx$ requires conditioning the joint mixture on $\rvz_{\gamma}$. In the following, we lay out the necessary steps.

Assume that the mixture of posterior-prior encodings is parameterized by $C$ Gaussians along with their weights $w^{(c)}$
\begin{equation}\label{eq:joint_mixture}
p(\rvz_{\phi\gamma\text{mix}})=\textstyle\sum^C_c w^{(c)} \mathcal{N}(\rvz_{\phi\gamma\text{mix}}; \vmu_{\phi\gamma\text{mix}}^{(c)}, \mSigma_{\phi\gamma\text{mix}}^{(c)})\ .
\end{equation}
The random variable realization $\rvz_{\phi\gamma\text{mix}}$ can be split into $\rvz_{\phi\gamma\text{mix}}{=}[\rvz_1;\rvz_2]$, $\text{dim}(\rvz_1){=}\text{dim}(\rvz_{\phi})$ and $\text{dim}(\rvz_2){=}\text{dim}(\rvz_{\gamma})$. Note that $\rvz_1$ and $\rvz_2$ ``belong" to $p(\rvz_{\phi\gamma\text{mix}})$ and are not the same as $\rvz_\phi$ and $\rvz_\gamma$. Thus, each component $c$ is factored as
\begin{equation}
\vmu_{\phi\gamma\text{mix}}^{(c)} = 
\begin{bmatrix}
\vmu^{(c)}_1 \\ 
\vmu^{(c)}_2
\end{bmatrix}, \quad
\mSigma_{\phi\gamma\text{mix}}^{(c)} = 
\begin{bmatrix}
\mSigma^{(c)}_{11} & \mSigma^{(c)}_{12} \\ 
\mSigma^{(c)}_{21} & \mSigma^{(c)}_{22}
\end{bmatrix}\ .
\end{equation}
The aim is to compute the conditional mixture distribution $p(\rvz_1|\rvz_2){=}\textstyle\frac{p(\rvz_1,\rvz_2)}{p(\rvz_2)}$. In~\cite{bishop2006pattern}, the conditional distribution of a component in a multivariate Gaussian is given by
\begin{align}
&\mathcal{N}(\rvz_1|\rvz_2; \vmu_{\phi\gamma\text{mix}}^{(c)}, \mSigma_{\phi\gamma\text{mix}}^{(c)})=\mathcal{N}(\rvz_1;\vmu^{(c)}_{1|2}, \mSigma^{(c)}_{1|2})\ ,\\
&\vmu^{(c)}_{1|2}=\vmu^{(c)}_1 + \mSigma^{(c)}_{12}(\mSigma^{(c)}_{22})^{-1}(\rvz_2 - \vmu^{(c)}_2)\ ,\\
&\mSigma^{(c)}_{1|2}=\mSigma^{(c)}_{11} - \mSigma^{(c)}_{12}(\mSigma^{(c)}_{22})^{-1}\mSigma^{(c)}_{21}\ .
\end{align}
The marginal distribution $p(\rvz_2)$ is given simply by $\textstyle\sum^C_c w^{(c)} \mathcal{N}(\rvz_2; \vmu^{(c)}_{22}, \mSigma^{(c)}_{22})$. Thus, $p(\rvz_1|\rvz_2)$ is computed by
\begin{align}\label{eq:cond_mixture}
p(\rvz_1|\rvz_2){=}\textstyle \sum^C_c \frac{w^{(c)} \mathcal{N}(\rvz_2; \vmu^{(c)}_{22}, \mSigma^{(c)}_{22})}{p(\rvz_2)} \mathcal{N}(\rvz_1;\vmu^{(c)}_{1|2}, \mSigma^{(c)}_{1|2})\ ,
\end{align}
where the fraction provides the new mixture weights that are normalized by the density of the marginal $p(\rvz_2)$. In this manner, we can sample $\rvz_{\text{cond-mix}}{\sim}p(\rvz_1|\rvz_2{=}\rvz_\gamma)$ and feed the decoder with $\rvz_{\text{cond-mix}}$ instead of $\rvz_{\gamma}$, $p_{\theta}(\rvy|\rvx,\rvz_{\text{cond-mix}})$. Thus, $p(\rvz_1|\rvz_2)$ serves as a link to obtain a latent vector close to what would be a posterior encoding (through their joint mixture relationship), considering that the posterior is not available in inference. See Fig. \ref{fig:cond-expost-sampling} (bottom) for an illustration.

The conditional mixture in~\eqref{eq:cond_mixture} provides a more expressive sampling distribution than the simplistic Gaussian prior. The conditioning by the prior sample results in a distribution that contains similar ground-truth training posteriors. Further, the weights of the conditional mixture (the fraction in ~\eqref{eq:cond_mixture}), different to $w^{(c)}$ in \eqref{eq:joint_mixture}, can effectively prune irrelevant components by assigning low values, potentially providing ${\ll}C$ components with non-negligible weights. In this way, a variable number of components in the latent space can be modeled based on the encountered context.

In the context of the CXP-\ac{CUAE} model, an open question is choosing the specific $\rvz_{\gamma}$ vector to condition the joint mixture. Considering that the prior $\gamma$ can provide sigma points, each of the $2n{+}1$ distinct points would result in a different conditional mixture. We choose the sigma point that incurs the largest density in the marginal distribution $p(\rvz_2)$ (the denominator term in~\eqref{eq:cond_mixture}). Intuitively, such a sigma point would cut the joint mixture where it is most data-rich.

\subsection{Output Trajectory Generation}\label{subsec:traj_gen}
\begin{table*}[t!]
    \centering
            \adjustbox{max width=\textwidth}{
        \begin{tabular}{lccccccccccccccc}
            \toprule
                \multirow{2}{*}{Model}&
                \multirow{2}{*}{sampling}&
                \multirow{2}{*}{latent space}&
                \multirow{2}{*}{inference via}&
                \multirow{2}{*}{$K$}&
                \multirow{2}{*}{$M$}&
                \multirow{2}{*}{minADE}&
                \multirow{2}{*}{minFDE}&
                \multicolumn{3}{c}{mixture-NLL}&
                & 
                \multicolumn{3}{c}{winner-NLL}\\
            \cline{9-11}\cline{13-15}
            &&&&&&&&1s&2s&3s&&1s&2s&3s\\

            \midrule
            \midrule
            
            CVAE & random & Gaussian & latent space & 6 & 6 & 0.149 & 0.478 & - & - & - && 1.846 & 1.877 & 2.167 \\
            CVAE & random & Gaussian & latent space & 65 & 65 & 0.078 & 0.209 & - & - & - && 1.845 & 1.854 & 1.955 \\
            \midrule
            \midrule
            \rowcolor{Gainsboro!40}
            CUAE & unscented & Gaussian & latent space & 6 & 6 & 0.145 & 0.452 & - & - & - && 1.846 & 1.872 & 2.131 \\
            \rowcolor{Gainsboro!40}
            CUAE & unscented & Gaussian & latent space & 65 (all) & 65 & 0.087 & 0.170 & - & - & - && 1.845 & 1.846 & 1.899 \\
            \midrule
            \rowcolor{Gainsboro!40}
            CVAE\textit{+clusters} & random & Gaussian & latent space & 65 & 6 & 0.134 & 0.454 & 1.842 & 1.928 & 2.607 && -2.685 & -0.382 & 1.980 \\
            \rowcolor{Gainsboro!40}
            CUAE\textit{+clusters} & unscented & Gaussian & latent space & 65 (all) & 6 & 0.130 & 0.400 & 1.850 & 2.010 & 2.780 && \textbf{-2.701} & \textbf{-1.358} & \textbf{1.011} \\
            \midrule
            \rowcolor{Gainsboro!40}
            CXP-CVAE\textit{+clusters} & random & Gaussian & cond. ex-post & 65 & 6 & 0.128 & 0.427 & \textbf{-2.556} & 3.400 & 6.608 && - & - & - \\
            \rowcolor{Gainsboro!40}
            CXP-CUAE\textit{+clusters} & unscented & Gaussian & cond. ex-post & 65 (all) & 6 & 0.122 & 0.379 & -2.431 & 0.336 & 2.792 && - & - & -\\
            \midrule
            \rowcolor{Gainsboro!40}
            GMM-CVAE & random & GMM & latent space & 65 & 6 & 0.101 & 0.301 & -0.437 & \textbf{-0.016} & 1.152 && - & - & -\\
            \rowcolor{Gainsboro!40}
            GMM-CUAE & unscented & GMM & latent space & 65 (all) & 6 & \textbf{0.097} & \textbf{0.283} & -0.430 & 0.053 & \textbf{1.150}  && - & - & -\\
            \bottomrule
            \bottomrule
        \end{tabular}
        }

        \caption{\footnotesize{Left half: Breakdown of the approaches described in Sec.~\ref{sec:method}. Note that all of the approaches in the gray-colored rows below the \ac{CVAE} rows are proposed in this work. Right half: Corresponding INTERACTION experiment results. Legend: $K$ -- number of samples/sigmas from the latent space (6 or 65 due to a 32-dim. latent space), $M$ -- number of final provided trajectories, \textit{+clusters} -- $K$ decoded samples/sigmas in the output are clustered into $M$ trajectories, GMM latent space -- $M$ trajectories are given by the centroids of $K$ decoded samples/sigmas per each GMM component.}}
    \label{tab:pred_quant_results}
\vspace{-15pt}
\end{table*}
Commonly used metrics such as \ac{minADE} and \ac{minFDE} (see~\cite{knittel2023dipa} for definitions) necessitate a fixed number of $M$ candidate trajectories, e.g. $M{=}6$. \ac{CVAE}s inherently exhibit large variance on such metrics due to the random sampling. Even the deterministic sampling of the \ac{CUAE} poses the question of which $M$ sigmas to provide among $2n{+}1$ choices, $M{\ll}2n{+}1$. Therefore, we investigate a simple way to provide a more consistent output. We first draw a large number of $K$ random samples (\ac{CVAE}) or take all $K{=}2n{+}1$ latent sigmas (\ac{CUAE}). Then, we cluster them into $M$ clusters with a k-means procedure and provide only the centroids. A similar approach is explored in~\cite{deo2022multimodal}. Thus, we also evaluate the clustering-enhanced \ac{CVAE}, \ac{CUAE}, and the \ac{CXP}-\ac{CUAE}, detailed in Sec.~\ref{subsubsec:cvae},~\ref{subsubsec:cuae}, and~\ref{subsubsec:cond-expost}, respectively. We do not apply it to the latent space \ac{GMM} from Sec~\ref{subsubsec:mixture_cvae} since it already has a mechanism to provide fixed $M$ trajectories through the $C{=}M$ components. We expect that this approach especially boosts the performance of the \ac{CUAE} in training, since it translates its structured latent space coverage into the output space. Overall, as we offer multiple models touching different facets of the \ac{CVAE}, we summarize our proposed approaches in Tab.~\ref{tab:pred_quant_results} (left). 

\vspace{-1pt}
\section{Results}
\vspace{-1pt}
Here, we describe the experimental setup and present the results of our proposed \ac{CVAE} trajectory prediction models. As \ac{CVAE}s are used beyond this task, our architectural improvements are not limited to prediction, which is the primary evaluation setting. Thus, to better understand our models, we extend the evaluation with the secondary setting of classical image modeling on the rich CelebA~\cite{liu2015faceattributes} dataset. 
\vspace{-10pt}
\subsection{Implementation}
\vspace{-1pt}
The network architectures of our \ac{CVAE} approaches explicitly follow the StarNet model~\cite{janjovs2022starnet}. It is a deterministic, single-agent\footnote{Our \ac{CVAE}-level improvements have no inherent restrictions toward a joint prediction extension, which is a more sound approach to the problem.} predictor that uses a graph-based map and trajectory history context. We use a shared StarNet encoder, comprising a 1D-CNN trajectory history network, GNN map network and an attention-based~\cite{vaswani2017attention} agent interaction network as a basis for the posterior $\phi$ and prior $\gamma$ distributions. Since the posterior $\phi$ additionally receives the ground-truth future trajectory, we reuse the 1D-CNN. Thus, in both $\phi$ and $\gamma$ the StarNet encoder produces a single feature vector passed onto a two-layer $[128, 128]$ MLP with batch normalization and ReLU activation. The output of this MLP is passed onto two separate 32-dim. layers producing $\vmu_\phi$ or $\vmu_\gamma$ and $\log\bm{\sigma}^2_\phi$ or $\log\bm{\sigma}^2_\gamma$ (used to construct diagonal covariance matrices). In the \ac{GMM}-\ac{CVAE}\footnote{The term \ac{GMM}-\ac{CVAE} was previously used in~\cite{hong2019rules} in a prediction context and in~\cite{wang2017diverse} in image modeling. The former uses a categorical latent space and regresses an output-level \ac{GMM}. The latter uses pre-defined clusters as modes of the \ac{GMM}, where cluster labels are already available instead of being learned in the training process. Since both approaches significantly differ from ours, we reuse the term for our latent-space \ac{GMM}.} in Sec.~\ref{subsubsec:mixture_cvae}, an additional 64-dim. layer produces weights $w_\phi$ or $w_\gamma$ from concatenated means and variances as input. The StarNet decoder $\theta$ predicts future trajectories in an action-based manner\footnote{The model first predicts future actions (acceleration and steering angle) and then unrolls them into future positions (starting from the current position) using a kinematic bicycle model.}~\cite{janjovs2021self}. We emphasize that other more sophisticated models can be used within the encoder/decoder, which is orthogonal to our top-level \ac{CVAE}.

In image modeling experiments, we use the identical setup from~\cite{janjovs2023unscented} and extend it with a prior $\gamma$ network. It encodes the conditioning $\rvx$ in the CelebA dataset consisting of a 40-dim. binary vector of face attributes (whereas the ground-truth $\rvy$ acc. to Fig.~\ref{fig:cvae} is an image). The prior $\gamma$ is realized as a $[64, 64]$ MLP for both $\vmu_\gamma$ and $\log\bm{\sigma}^2_\gamma$ with a shared first layer.

\subsection{Datasets and Training Setup}
We trained and evaluated our trajectory prediction models on the INTERACTION~\cite{zhan2019interaction} dataset of highly interactive driving containing merges, roundabouts, and intersections. We used the official training and validation splits, predicting 3s trajectories ($T{=}30$ at 10Hz) given a 1s history. In image modeling, we used the rich CelebA dataset~\cite{liu2015faceattributes} of human faces, containing $64{\times}64{\times}3$ images pre-processed the same way as in~\cite{janjovs2023unscented} and 40-dim. binary attribute annotations\footnote{Examples include: \textit{Smiling}, \textit{Eyeglasses}, \textit{Young}, \textit{Blond\_Hair}.}.

The prediction models are trained for 30 epochs with Adam~\cite{kingma2014adam}, starting from a $1e^{-4}$ learning rate and halving it for epochs 10, 15, 20, and 25. CelebA experiments use the same setup as in~\cite{janjovs2023unscented}: 100 epochs and a learning rate halving on loss plateau. All models are implemented in PyTorch~\cite{paszke2019pytorch}. In \ac{CXP} estimation, we used~\cite{borchert2022pycave} (compatible with~\cite{scikit-learn}) for fast GPU-based \ac{GMM} fitting after training, taking around 30 min. over the entire training set. In both use-cases, the models took around 1.5 days to train on a single Nvidia 3090 GPU.\looseness=-1

\subsection{Image Modeling Performance}
In this secondary evaluation setting, our aim is to assess the proposed models' ability to generate realistic images. One goal is reconstructing existing images by compressing and decompressing them from the latent space (a task trajectory prediction models are not evaluated on). More specifically, a trained \ac{CVAE} encodes a ground-truth image $\rvy$ and its attributes as context $\rvx$ into the posterior distribution $\mathcal{N}({\rvz_{\phi}|\rvx,\rvy;\vmu_\phi, \mSigma_\phi})$. Then, it feeds a random sample (or sigma) $\rvz_{\phi,k}$ to the decoder, which reconstructs the output image $\rvy_k{=}f_\theta(\rvx, \rvz_{\phi,k})$. This process is conceptually the same as the \ac{CVAE} in Fig.~\ref{fig:cvae}. Furthermore, we evaluate the ability to generate realistic new image samples. In this manner, the model only receives the context $\rvx$ encoded into the prior $\mathcal{N}({\rvz_{\gamma}|\rvx;\vmu_\gamma, \mSigma_\gamma})$. Then, the decoder produces an image $\rvy_k{=}f_\theta(\rvx, \rvz_{\gamma,k})$ using a sample $\rvz_{\gamma,k}$.  In contrast, the \ac{CXP}-\ac{CVAE} produces an image in inference using a sample from the conditional mixture, $\rvy_k{=}f_\theta(\rvx, \rvz_\text{cond-mix})$. For all models, we use four samples (or random sigmas) in training. In both reconstruction and sampling, we evaluate image realism with the established \ac{FID}~\cite{heusel2017gans}, which computes the Wasserstein metric between sets of real and sampled images.

Tab.~\ref{tab:celeba_results} shows quantitative results comparing the vanilla \ac{CVAE}, \ac{CUAE}, and both with \ac{CXP} estimation ($C{=}10$, see \eqref{eq:joint_mixture}). We do not include GMM-\ac{CVAE} since generating multiple image outputs is usually not relevant to the problem (which is stationary). %
We additionally ablate the baseline \ac{XP} estimation from~\cite{janjovs2023unscented},~\cite{ghosh2020variational}, which does not condition on attributes $\rvx$; it fits the mixture only on $\rvz_{\phi,k}$ and directly samples $\rvz_{\phi\text{mix}}$ to feed the decoder $\rvy_k{=}f_\theta(\rvx, \rvz_{\phi\text{mix}})$. We observe that the best scores are achieved by \ac{CUAE} models with (C)XP estimation. Fig.~\ref{fig:qualitative_results_celeba} qualitatively corroborates the results from Tab.~\ref{tab:celeba_results}; it is evident that such models generate sharper and more realistic images than prior inference models. As expected though, the \ac{XP}-\ac{CVAE} struggles to include the queried attribute into the image (since the sample $\rvz_{\phi\text{mix}}$ does not contain it), something vanilla \ac{CVAE} and our \ac{CXP}-\ac{CVAE} are well capable of.
\def\facew{0.44}
\renewcommand\arraystretch{0.25}
\begin{figure}[t!]
	\centering
	\resizebox{1\columnwidth}{!}{%
		\begin{tabularx}{1\columnwidth}{tss}
			& \footnotesize Reconstruction & \footnotesize Sampling (given attribute) \\
			\toprule
			\scriptsize \raggedleft GT & \includegraphics[width=\facew\columnwidth]{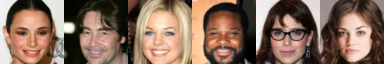} & \hspace{0.1cm} \scriptsize \textit{bh} \hspace{0.32cm} \scriptsize \textit{b} \hspace{0.29cm} \scriptsize \textit{we} \hspace{0.22cm} \scriptsize \textit{wh} \hspace{0.22cm} \scriptsize \textit{m} \hspace{0.25cm} \scriptsize \textit{gh} \\
            \scriptsize \raggedleft CVAE & \includegraphics[width=\facew\columnwidth]{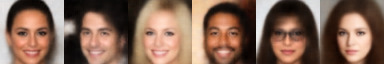} & \includegraphics[width=\facew\columnwidth]{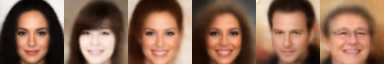}
            \\
			\scriptsize \raggedleft XP-\ac{CUAE} & \includegraphics[width=\facew\columnwidth]{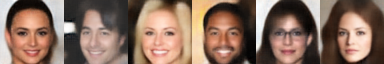} & \includegraphics[width=\facew\columnwidth]{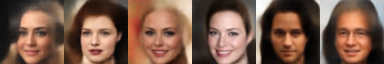} \\
			\scriptsize \raggedleft CXP-\ac{CUAE} & \includegraphics[width=\facew\columnwidth]{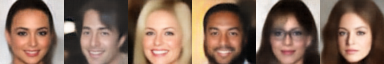} & \includegraphics[width=\facew\columnwidth]{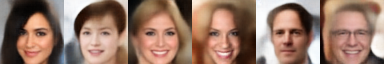} \\
			\bottomrule
	       \end{tabularx}
    }
	\caption{\footnotesize{Example reconstructed and sampled images (best viewed in color and zoomed in). The \ac{XP}-\ac{CUAE} and \ac{CXP}-\ac{CUAE} inference models generate sharper images than vanilla \ac{CVAE}, as also evidenced by the quantitative results of corresponding rows in Tab.~\ref{tab:celeba_results}. Furthermore, the proposed \ac{CXP} inference succeeds at incorporating the attributes into samples, while the baseline \ac{XP} inference struggles with it. Attributes: \textit{Black\_Hair} (\textit{bh}), \textit{Bangs} (\textit{b}), \textit{Wearing\_Earrings} (\textit{we}), \textit{Wavy\_Hair} (\textit{wh}), \textit{Male} (\textit{m}), \textit{Gray\_Hair} (\textit{gh}).}}
	\label{fig:qualitative_results_celeba}
\vspace{-15pt}
\end{figure}
\renewcommand\arraystretch{1}

\renewcommand\arraystretch{0.7}
\begin{table}[t!]
	\centering
    \adjustbox{max width=\textwidth}{
        \begin{tabular}{lcccc}
            \toprule
                \scriptsize \multirow{2}{*}{Model}&
                \scriptsize \multirow{2}{*}{sampling}&
                \scriptsize \multirow{2}{*}{inference via}&
                \multicolumn{2}{c}{\scriptsize image modeling}\\
            \cline{4-5}
            & & & \scriptsize reconstr. & \scriptsize sampling \\
            \midrule
            \scriptsize CVAE & \scriptsize random & \scriptsize latent space & \scriptsize 59.74 & \scriptsize 62.61 \\
            \scriptsize CUAE & \scriptsize unscented & \scriptsize latent space & \scriptsize 47.92 & \scriptsize 98.50 \\
            \midrule
            \scriptsize XP-CVAE & \scriptsize random & \scriptsize ex-post & \scriptsize 59.29 & \scriptsize 63.70 \\
            \scriptsize XP-CUAE & \scriptsize unscented & \scriptsize ex-post & \scriptsize 40.67 & \scriptsize 48.83\\
            \midrule
            \scriptsize CXP-CVAE & \scriptsize random & \scriptsize cond. ex-post & \scriptsize 59.32 & \scriptsize 63.53 \\
            \scriptsize CXP-CUAE & \scriptsize unscented & \scriptsize cond. ex-post & \scriptsize \textbf{40.44} & \scriptsize \textbf{48.52} \\
            \bottomrule
        \end{tabular}
}
  
    \caption{\footnotesize{Quantitative image modeling results on \ac{FID} (lower is better). We ablate the proposed unscented sampling, ex-post estimation (XP)~\cite{ghosh2020variational,janjovs2023unscented}, and the proposed conditional ex-post estimation (CXP).}}
	\label{tab:celeba_results}
\vspace{-10pt}
\end{table}
\renewcommand\arraystretch{1}

\renewcommand\arraystretch{0.7}
\begin{table}[t!]
	\centering
    \adjustbox{max width=\textwidth}{
\begin{tabular}{lcccc}
	\toprule
	
    \scriptsize Model & \scriptsize minADE$_6$ & \scriptsize minFDE$_6$\\
	\midrule
	\scriptsize ITRA \cite{scibior2021imagining}  &  \scriptsize 0.17 & \scriptsize 0.49 \\
	\scriptsize GOHOME \cite{gilles2022gohome} & \scriptsize - & \scriptsize 0.45  \\
    \scriptsize joint-StarNet \cite{janjovs2022starnet} &\scriptsize  0.13 & \scriptsize 0.38 \\
	\scriptsize DiPA \cite{knittel2023dipa} & \scriptsize 0.11 & \scriptsize 0.34  \\
    \scriptsize MB-SS-ASP \cite{janjovs2023bridging} & \scriptsize 0.10 & \scriptsize 0.30 \\
    \scriptsize SAN \cite{janjovs2022san} & \scriptsize 0.10 & \scriptsize 0.29 \\
	\cmidrule(l){2-3} 
	\scriptsize GMM-CUAE & \scriptsize \textbf{0.097} & \scriptsize \textbf{0.283} \\
	\bottomrule
\end{tabular}}
  
    \caption{\footnotesize{Comparison of the best model in Tab.~\ref{tab:pred_quant_results} with models from the literature on the INTERACTION validation dataset.}}
	\label{tab:prediction_literature}
\vspace{-5pt}
\end{table}
\renewcommand\arraystretch{1}

\begin{figure}[t!]
\centering
\begin{subfigure}{0.24\columnwidth}
\includegraphics[width=1\textwidth,trim={1cm 1cm 2cm 0},clip]{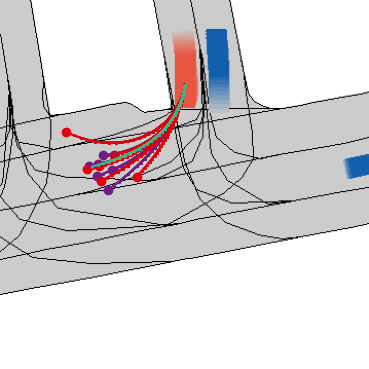}
\end{subfigure}\hspace{-0.1cm}
\begin{subfigure}{0.24\columnwidth}
\includegraphics[width=1\textwidth,trim={1.5cm 1cm 1.5cm 0},clip]{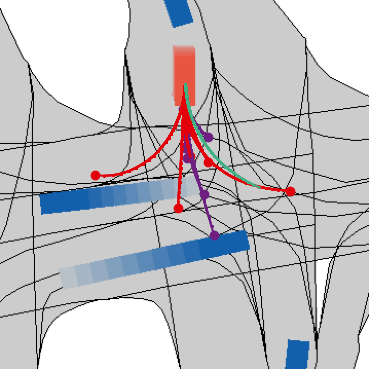}
\end{subfigure}\hspace{-0.1cm}
\begin{subfigure}{0.24\columnwidth}
\includegraphics[width=1\textwidth,trim={2cm 1cm 1cm 0},clip]{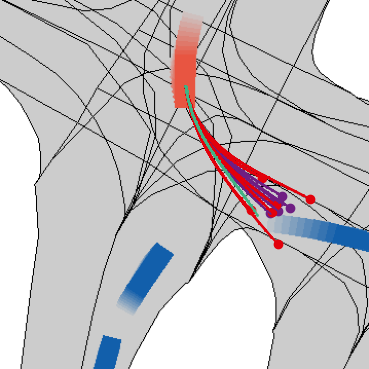}
\end{subfigure}\hspace{-0.1cm}
\begin{subfigure}{0.24\columnwidth}
\includegraphics[width=1\textwidth,trim={1cm 0.3cm 2cm 0.7cm},clip]{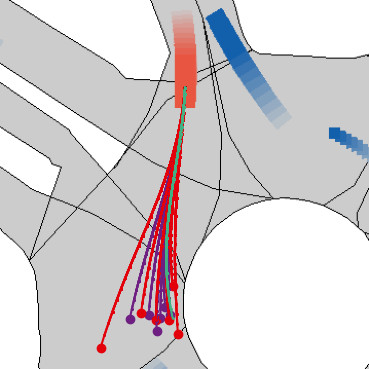}
\end{subfigure}
\caption{\footnotesize{Qualitative comparison of sampling choices (best viewed in color): trajectories reconstructed from sigma points (red) are significantly more diverse than random samples (purple). Traffic participants are depicted in blue and the predicted vehicle in red, with fading for history.}}
\label{fig:prediction_qual_results}
\vspace{-20pt}
\end{figure}

\subsection{Trajectory Prediction Performance}
In this section, we pit our proposed \ac{CVAE} improvements against each other: \ac{CUAE} (Sec.~\ref{subsubsec:cuae}), \ac{GMM} latent space (Sec.~\ref{subsubsec:mixture_cvae}), \ac{CXP} (Sec.~\ref{subsubsec:cond-expost}), as well as output-level clustering (Sec.~\ref{subsec:traj_gen}) on the primary, trajectory prediction evaluation setting. We aim to evaluate the quality of multi-modal predictions on a trajectory and distribution level using the \ac{minADE} and \ac{minFDE} metrics as well as the distributional \ac{NLL}. Comprehensive results are shown in Tab.~\ref{tab:pred_quant_results} (right). Since our approaches relate to core \ac{CVAE} aspects, our main baseline is a vanilla \ac{CVAE}, however, Tab.~\ref{tab:prediction_literature} shows a comparison of our highest-performing model with non-\ac{CVAE} approaches in literature.\looseness=-1

The first four rows of Tab.~\ref{tab:pred_quant_results} show the results of \ac{CVAE} and \ac{CUAE} models from Sec.~\ref{subsec:lat_sampling}. We trained with $\sigma{=}1.0$ in \eqref{eq:rec-samples} and \eqref{eq:rec-dist} (it has been shown to work well for sample-based predictors~\cite{dendorfer2021mg}). Since it is nontrivial to compute the full \ac{NLL}, we compute it only for the closest mode to the ground truth ($\sigma{=}1.0$). We see that the \ac{CUAE} provides a 5\% boost over \ac{CVAE} in trajectory metrics, however, qualitative results in Fig.~\ref{fig:prediction_qual_results} show the potential of sigma points to provide good coverage, as illustrated in Fig.~\ref{fig:sampling-illustration}. Further, both \ac{CVAE} and \ac{CUAE} benefit from increasing the number of samples or using all sigma points (32-dim. latent space yields $K{=}65$), though these models are not comparable with the rest. The output clustering from Sec.~\ref{subsec:traj_gen} provides $M{=}6$ trajectories based on the $K$ samples/sigmas. It especially benefits the \ac{CUAE}; its thorough output-space coverage is encapsulated into few higher quality candidates than the \ac{CVAE}'s clustered outputs. The {\textgreater}50\% lower winner-NLL (only the cluster centroid closest to the ground truth is evaluated) shows that clusters on sigma trajectories are more meaningful. However the mixture-NLL, computed using ratios of members in each cluster as weights, is high. It shows that such weights are not a good proxy for the actual mixture distribution.\looseness=-1

\ac{CXP} provides an alternative to the latent prior in inference. We apply it to the clustering models in which we fit a $C{=}50$-component \ac{GMM} from the training set posterior and prior encodings after training. Then, we condition it as described in Sec.~\ref{subsubsec:cond-expost} and draw $K{=}65$ samples from the conditional mixture, clustered into $M{=}6$ trajectories. \ac{CXP} brings additional gains in trajectory metrics, showing that a more expressive inference distribution is beneficial. However, it does not provide an easy way to compute the mixture \ac{NLL}. We approximate it via the conditional mixture weights and component mean sigma trajectories. Finally, the GMM latent model (with $C{=}M{=}6$) provides the best scores, significantly outperforming the previous models in trajectory- and distribution-level metrics. It shows that inducing a more expressive latent space in a \ac{CVAE} context (while accordingly approximating the \ac{ELBO}) brings significant performance improvements. Here, random and unscented sampling perform similarly well. However, it is important to note that the \ac{GMM} reconstruction loss function in~\eqref{eq:covar_mm_reconst} already incorporates a crucial aspect of the \ac{UT}, as described in Sec.~\ref{subsubsec:cuae}. The reconstruction error is computed in a \ac{CUAE}-like manner (see \eqref{eq:rec-dist}), using the likelihood of the centroid trajectory of the winner \ac{GMM} component (where the centroid is computed either by transforming random samples or sigmas), rather than in a \ac{CVAE}-like manner (see \eqref{eq:rec-samples}), using the expectation of individually reconstructed trajectories.

Overall, the obtained results comprehensively demonstrate the efficacy of our proposed alternatives to key components of the trajectory prediction \ac{CVAE}, answering the research questions \hyperlink{rq1}{\textit{(i)}} and \hyperlink{rq2}{\textit{(ii)}} posed in Sec.~\ref{sec:intro}. First, the unscented sampling and latent space transformation can bring greater diversity in output trajectories while outperforming the ubiquitous random sampling. Second, alternative inference procedures, validated in non-prediction contexts as well, can utilize substantially more expressive distributions to draw higher-quality samples, while requiring no training overhead. Finally, greatest advantages can be found in directly training a more expressive Gaussian mixture latent space whose components map to a structured multi-modal output distribution.

\section{Conclusion}
In this paper, we investigated important shortcomings of the \ac{CVAE} in trajectory prediction. We answered questions surrounding latent space assumptions by showing that unscented sampling and mixture models in training and inference provide high performance alternatives to existing structures. We anticipate that our findings will lead to a more effective usage of \ac{CVAE} models in prediction and beyond.

\bibliographystyle{IEEEtran}
\bibliography{IEEEabrv,references}

\end{document}